%% file: TECsubmit.tex
\documentclass[dvipsnames,format=sigconf,anonymous=false,review=false]{acmart}

\setcopyright{none} 
\acmDOI{} 
\acmISBN{} 
\acmYear{2025} 
\copyrightyear{2025} 
\acmConference[GECCO 2025]{the Genetic and Evolutionary Computation Conference}{This is the authors' version of the paper accepted at GECCO 2025.}

\usepackage{fancyhdr}
\usepackage[T1]{fontenc}
\usepackage{color}
\usepackage{amsmath}
\usepackage{amsthm}
\usepackage{url}
\usepackage{graphicx}
\usepackage{algorithm2e}
\usepackage{algorithmic}
\usepackage{psfrag}
\usepackage{subcaption}
\usepackage{tabularx}

\newtheorem{theorem}{Theorem}

\newtheorem{observation}{Observation}

\begin{document}

\newcommand{\wek}[1]{
	{\bf #1} 
}
\newcommand{\mat}[1]{
	{\bf #1} 
}
\newcommand{\srednia}[1]{
	\overline{ #1 } 
}

\newcommand*{\tran}{{\mkern-1.5mu\mathsf{T}}}

\newcommand\norm[1]{\left\lVert#1\right\rVert}

\title{Covariance Matrix Adaptation Evolution Strategy without a matrix}

\author{Jaros{\l}aw Arabas}
\orcid{0000-0002-5699-947X}
\affiliation{\institution{Warsaw University of Technology} \department{Institute of Computer Science}
\city{Warsaw}
\country{Poland}}
\email{jaroslaw.arabas@pw.edu.pl}

\author{Adam Stelmaszczyk}
\orcid{0009-0009-0434-3434}
\affiliation{\institution{Warsaw University of Technology} \department{Institute of Computer Science}
\city{Warsaw}
\country{Poland}}
\email{adam.stelmaszczyk@pw.edu.pl}

\author{Eryk Warchulski}
\orcid{0000-0003-1416-7031}
\affiliation{\institution{Warsaw University of Technology} \department{Institute of Computer Science}
\city{Warsaw}
\country{Poland}}
\email{eryk.warchulski@pw.edu.pl}

\author{Dariusz Jagodzi\'nski}
\orcid{0000-0001-7938-5916}
\affiliation{\institution{Warsaw University of Technology} \department{Institute of Computer Science}
\city{Warsaw}
\country{Poland}}
\email{dariusz.jagodzinski@pw.edu.pl}

\author{Rafa{\l} Biedrzycki}
\orcid{0000-0001-6730-4269}
\affiliation{\institution{Warsaw University of Technology} \department{Institute of Computer Science}
\city{Warsaw}
\country{Poland}}
\email{rafal.biedrzycki@pw.edu.pl}

\begin{abstract}
Covariance Matrix Adaptation Evolution Strategy (CMA-ES) is a highly effective optimization technique. A primary challenge when applying CMA-ES in high dimensionality is sampling from a multivariate normal distribution with an arbitrary covariance matrix, which involves its  decomposition. The cubic complexity of this process is the main obstacle to applying CMA-ES in high-dimensional spaces.

We introduce a version of CMA-ES that uses no covariance matrix at all. In the proposed matrix-free CMA-ES, an archive stores the vectors of differences between individuals and the midpoint, normalized by the step size. New individuals are generated as the weighted combinations of the vectors from the archive. We prove that the probability distribution of individuals generated by the proposed method is identical to that of the standard CMA-ES.

Experimental results show that reducing the archive size to store only a fixed number of the most recent populations is sufficient, without compromising optimization efficiency. The matrix-free and matrix-based CMA-ES achieve comparable results on the quadratic function when the step-size adaptation is turned off. When coupled with the step-size adaptation method, the matrix-free CMA-ES converges faster than the matrix-based, and usually yields the results of a comparable or superior quality, according to the results obtained for the CEC'2017 benchmark suite.

Presented approach simplifies the algorithm, offers a novel perspective on covariance matrix adaptation, and serves as a stepping stone toward even more efficient methods.

\end{abstract}


\maketitle

\section{Introduction}

For continuous optimization problems, the Covariance Matrix Adaptation Evolution Strategy (CMA-ES) \cite{hansen2006cma} stands as one of the leading methods within the stochastic algorithms family. CMA-ES adapts the covariance matrix of the search distribution, allowing it to learn and exploit the underlying structure of the objective function. This capability to navigate complex optimization landscapes has garnered significant attention across various domains, including scientific research, engineering and industrial applications.

Each iteration of CMA-ES involves generating new candidate solutions from a multivariate normal distribution. This distribution is dynamically updated based on the performance of the generated solutions. However, sampling from this distribution necessitates the decomposition of the covariance matrix, resulting in a computational complexity of $\mathcal{O}(n^3)$ per iteration, where $n$ is the number of dimensions. This cubic scaling presents a significant bottleneck when applying CMA-ES to high-dimensional problems involving thousands of variables. To address this computational challenge, several approaches have been proposed.

In \cite{SEPCMAES-2008} it was observed that the overall cost of running CMA-ES can be reduced by decomposing the covariance matrix not in every but, for example, every tenth iteration. According to the authors, this modification introduces only a small deterioration of the overall optimization efficiency. In another approach \cite{NIPS2016_CholeskyCMAES}, the authors propose to change the cumulative step-size adaption (CSA) mechanism in the CMA-ES and replace the inverse of the square root of the covariance matrix by the inverse of the triangular Cholesky factor. Despite these reductions, the matrix operations used in the CMA-ES algorithms for generating basis vectors or $\sigma$ step adaptation, remain expensive. 

Another approach to reducing the computational effort is to simplify the covariance matrix into a form that is more convenient for processing. In the MVA-ES method \cite{Poland2001}, the covariance matrix is defined as the sum of the identity matrix and the outer product of the cumulative midpoint shift. As a result, the process of generating difference vectors does not require matrix decomposition. While the MVA-ES method achieves results comparable to those of CMA-ES, this is true only when there is a single dominant direction for the adaptation of distribution parameters. Consequently, MVA-ES performs well for optimizing unimodal functions with a clear direction of steepest descent in the target values. However, for multimodal functions, the performance of MVA-ES is significantly worse than that of the baseline CMA-ES.

A similar approach to simplifying the covariance matrix, as in MVA-ES, is employed in the R1-NES algorithm \cite{Sun2013}. In this method, the covariance matrix is defined as the sum of the identity matrix and the outer product of a vector that represents the weighted average of recent gradients of the objective function. Consequently, the covariance matrix will have at most one eigenvector with a large eigenvalue. This limitation prevents the proposed procedure from effectively approximating the actual shape of the objective function. The study's results indicate that the R1-NES method is only effective for specific types of objective function shapes.

An alternative approach to reducing computational effort is to redefine the method for updating the covariance matrix. The MA-ES algorithm \cite{SMAES2017} eliminates the need for decomposition by approximating the square root of the covariance matrix, which is required to generate the population. Instead of accumulating midpoint shift vectors to update the covariance matrix, the method accumulates multivariate normal vectors that, after transformation into the search space, produced the best-fit points. These assumptions allow the algorithm to bypass matrix factorization without significantly compromising optimization efficiency. However, the proposed formula for the square root of the covariance matrix requires matrix multiplication in each generation. By employing the Coppersmith-Wino\-grad matrix multiplication method, the computational complexity of one iteration of the MA-ES algorithm can be reduced to $\mathcal{O}(n^{2.37})$.

The LM-MA-ES method \cite{Loshchilov3} is a modification of the MA-ES algorithm designed to further reduce computational effort. Instead of adapting a matrix that approximates the square root of the covariance matrix for subsequent multiplication by a realization of a unit normal random variable, the authors propose directly estimating and adapting the result of this multiplication. This approach reduces the computational complexity of the method to $\mathcal{O}(n \log(n))$. However, since LM-MA-ES approximates the basis vectors produced by the MA-ES algorithm, the resulting covariance matrix of the generated points may differ from that of the original CMA-ES algorithm.

A different approach to simplifying CMA-ES is presented in \cite{ArabasJ19}, where the authors propose an algorithm called the Differential Evolution Strategy. Rather than using Gaussian mutation to generate new individuals, this method is based on the differential mutation. New individuals are generated by adding difference vectors to the population midpoint, and each difference vector is a weighted combination of three elements: the difference vector between two randomly selected points from a randomly chosen past population, a cumulative midpoint shift observed in one randomly selected past generation, and a midpoint shift from a past population. These three vectors are combined using weights that are random values drawn from the standard normal distribution. The authors acknowledge that the resulting probability distribution differs from that of CMA-ES. In contrast, in this work, we propose a method whose distribution is equivalent to that of CMA-ES.

In most of the approaches mentioned above, the authors focus on addressing the consequences of aggregating the search history into a matrix. Rather than mitigating the consequences, it may be more effective to eliminate the root cause of the problem. To this end, we propose a solution where no matrix is needed at all.

We utilize an archive of populations to define new points by combining individuals with cumulated midpoint shift vectors, using randomly assigned weights for these combinations. The probability distribution of the generated points in our approach matches that of CMA-ES, i.e., the resulting distribution is multivariate normal with the same mean vector and covariance matrix as in the case of CMA-ES. It is important to emphasize that our motivation was to develop a method that is compliant with, rather than superior in terms of convergence, to the matrix-based CMA-ES.

The resulting algorithm cannot be coupled with the Cumulative Step-Size Adaptation (CSA) rule for adapting the mutation step size. Therefore, we employ the Previous Population Midpoint Fitness (PPMF) rule \cite{Warchulski2021} for step-size adaptation. When used alongside the matrix-based CMA-ES, PPMF has demonstrated competitiveness with CSA, particularly in optimization problems with a high number of dimensions.

The paper is organized as follows. In Section 2, we introduce the Vanilla CMA-ES method, outlining its core functionality. Section 3 presents the matrix-free CMA-ES (MF-CMA-ES), where we discuss the modification to CMA-ES that eliminates the need for matrix decomposition while maintaining the same statistical properties. Section 4 compares the performance of matrix-based and matrix-free CMA-ES under conditions where step-size adaptation is not applied. Section 5 extends the discussion by including step-size adaptation. Section 6 provides concluding remarks and suggestions for future work.

\section{Vanilla CMA-ES}

The starting point for our considerations is the vanilla  CMA-ES method \cite{hansen2006cma} --- see Fig.~\ref{alg:cmaes}. 
\begin{figure}[h!]
\begin{algorithmic}[1]
\STATE $\wek{p}_c^{(1)} \gets \wek{0}$, 
$\mat{C}^{(1)} \gets \mat{I}$
\STATE $t \gets 1$
\STATE initialize$(\wek{m}^{(1)},\sigma^{(1)})$
\WHILE{!stop}
  \STATE decompose $\mat{C}^{(t)}$ : $(\mat{L}^{(t)})(\mat{L}^{(t)})^\tran=\mat{C}^{(t)}$
   \FOR{$i=1$ \TO $\lambda$}
      \STATE $ \wek{z}_i^{(t)} \sim {\mathcal N}(\wek{0}, \mat{I}) $
      \STATE $ \wek{d}_i^{(t)} =\mat{L}^{(t)} \wek{z}_i^{(t)}$
      \STATE $ \wek{x}_i^{(t)} = \wek{m}^{(t)} + \sigma^{(t)} \wek{d}_i^{(t)} $
   \ENDFOR
   \STATE evaluate $(X^{(t)})$
   \STATE sort $(X^{(t)})$ according to fitness
   \STATE reorder $(D^{(t)})$ and $(Z^{(t)})$ according to $(X^{(t)})$   
   \STATE $\wek{\Delta}^{(t)} \gets \sum_{i=1}^\mu w_i \wek{d}_i^{(t)} $
   \STATE $\wek{m}^{(t+1)} \gets \wek{m}^{(t)} + \sigma^{(t)} \wek{\Delta}^{(t)}$
   \STATE $\wek{p}_c^{(t+1)} \gets (1-c_c)\wek{p}_c^{(t)} + \sqrt{\mu_\text{eff} c_c(2-c_c)} \cdot \wek{\Delta}^{(t)}$
   \STATE $\mat{C}_1^{(t)}= \wek{p}_c^{(t)}\left(\wek{p}_c^{(t)}\right)^\tran$
   \STATE $\mat{C}_\mu^{(t)}= \sum_{i=1}^\mu w_i\wek{d}_i^{(t)}\left(\wek{d}_i^{(t)}\right)^\tran$
   \STATE $\mat{C}^{(t+1)} \gets (1-c_{cov})\mat{C}^{(t)} + c_1 \mat{C}^{(t)}_1 + c_\mu  \mat{C}^{(t)}_\mu$
   \STATE $\sigma^{(t+1)} \gets $ update $(\sigma^{(t)} )   $
      
   \STATE $t \gets t+1$
\ENDWHILE

\end{algorithmic}
\caption{Outline of the matrix-based CMA-ES}
\label{alg:cmaes}
\end{figure}
CMA-ES operates by maintaining a multivariate normal distribution to sample candidate solutions during the optimization process. This distribution is parameterized by three components: the mean vector $\wek{m}^{(t)}$, the covariance matrix $\mat{C}^{(t)}$, and the step-size $\sigma^{(t)}$, where $t$ represents the iteration index. Initially, $\mat{C}^{(1)}$ is set to the identity matrix $\mat{I}$, while $\wek{m}^{(1)}$ and $\sigma^{(1)}$ are specified by the user.

At each iteration, the covariance matrix $\mat{C}^{(t)}$ is decomposed into its lower triangular form, $\mat{L}^{(t)}$, such that $\mat{L}^{(t)} (\mat{L}^{(t)})^\tran = \mat{C}^{(t)}$. 
The recommended approach for decomposing the covariance matrix is eigendecomposition \cite{hansen2006cma} since it works properly even when the matrix $\mat{C}^{(t)}$ is ill-conditioned. 

Using $\mat{L}^{(t)}$, the algorithm generates a population of $\lambda$ candidate solutions by first sampling $\lambda$ independent standard normal vectors $\wek{z}_i^{(t)} \sim \mathcal{N}(\wek{0}, \mat{I})$. These vectors are then transformed into difference vectors $\wek{d}_i^{(t)} = \mat{L}^{(t)} \wek{z}_i^{(t)}$. Finally, the candidate solutions, or individuals, are computed as $\wek{x}_i^{(t)} = \wek{m}^{(t)} + \sigma^{(t)} \wek{d}_i^{(t)}$.

Once the population is created, the candidates are sorted based on their fitness and the best $\mu$ individuals are selected. The difference vectors $\wek{d}_i^{(t)}$ and normal vectors $\wek{z}_i^{(t)}$ are reordered in compliance with $\wek{x}_i^{(t)}$. The difference vectors, which correspond to $\mu$ point with the best fitness, are weighted and aggregated to update the parameters of the distribution for the next iteration. Specifically, the mean vector $\wek{m}^{(t)}$ is adjusted using $\wek{\Delta}^{(t)}$ --- a weighted sum of the top $\mu$ difference vectors, scaled by the step-size $\sigma^{(t)}$. 
Value of $\wek{\Delta}^{(t)}$ is accumulated into the so called evolution path vector $\wek{p}_c^{(t+1)}$. Value of $\mu_\text{eff}$ is defined as $\mu_\text{eff}=\left(\sum_{i=1}^\mu (w_i)^2\right)^{-1}$.

The covariance matrix $\mat{C}^{(t)}$ is updated using a weighted combination of three components: the existing covariance matrix, a rank-1 update matrix $\mat{C}_1^{(t)}$ derived from the evolution path $\wek{p}_c^{(t)}$, and a rank-$\mu$ update matrix $\mat{C}_\mu^{(t)}$ formed from the top $\mu$ difference vectors.

The update rule for $\mat{C}^{(t)}$ is controlled by coefficients $c_1$, $c_\mu$, and $c_{cov}$, which satisfy the relations $c_{cov} = c_1 + c_\mu$ and $0 \leq c_1, c_\mu, c_{cov} \leq 1$.

\section{Matrix-free CMA-ES (MF-CMA-ES)}

According to the formula in Fig. \ref{alg:cmaes}, line 19, the covariance matrix $\mat{C}^{(t)}$ memorizes the history of populations of difference vectors $\wek{d}_i^{(t)}$ as well as the history of the difference vectors mean $\wek{\Delta}^{(t)}$. The formula has the form of a recursive expression that implements the exponential smoothing. Following the idea presented in \cite{ArabasJ19}, we substitute  the recursive formula that defines the covariance matrix with a non-recursive version.

\begin{observation}
    
The covariance matrix update rule (Fig. \ref{alg:cmaes}, line 19) can be expressed as:
\begin{align}
\mat{C}^{(t+1)}=&\sum_{\tau=1}^{t}(1-c_{cov})^{t-\tau} \left(c_1 \mat{C}_1^{(\tau)} + c_\mu \mat{C}_\mu^{(\tau)}\right)
+(1-c_{cov})^{t} \mat{I}
\label{eq:10}
\end{align}
where $\mat{C}_1$ and $\mat{C}_\mu$ are defined in Fig. \ref{alg:cmaes}, lines 17 and 18, respectively.
\end{observation}

Next, we formulate the following theorem.

\begin{theorem}
Consider the random vector:
\begin{align}
\wek{\delta}^{(t+1)}=&\sum_{\tau=1}^{t}(1-c_{cov})^{\frac{t-\tau}{2}} \left(c_\mu^{\frac{1}{2}} \sum_{j=1}^\mu w_j^{\frac{1}{2}} \wek{d}_j^{(\tau)}  {\mathcal N}(0,1) + c_1^{\frac{1}{2}} \wek{p}_c^{(\tau)} {\mathcal N}(0,1)\right) \nonumber \\
&+(1-c_{cov})^{\frac{t}{2}} {\mathcal N}(\wek{0},\mat{I})
\label{eq:th1}
\end{align}
where $\wek{d}_j^{(t)}$ and $\wek{p}_c^{(t)}$ are defined in Fig. \ref{alg:cmaes}, lines 8 and 16, respectively. Vectors $\wek{d}_j^{(t)}$ are ordered according to the fitness of their corresponding points $\wek{x}_j^{(t)}$. Symbols ${\mathcal N}(0,1)$ and ${\mathcal N}(\wek{0},\mat{I})$ stand for the mutually independent standard normal variates in one and in $n$ dimensions, respectively.

Probability distribution of the vector $\wek{\delta}^{(t+1)}$ is Gaussian with the covariance matrix equal $\Sigma[
\wek{\delta}^{(t+1)}]=\mat{C}^{(t+1)}$, where $\mat{C}^{(t+1)}$ is defined by the CMA-ES algorithm (Fig. \ref{alg:cmaes}, line 19).

\end{theorem}

\paragraph{Sketch of proof:}

The vector $\wek{\delta}^{(t+1)}$ is a linear combination of mutually independent multivariate normal variables, therefore it is multivariate normal.

Expected values of summands defining $\wek{\delta}^{(t+1)}$ equal
\begin{align}
E\left[\wek{d}_j^{(\tau)}  {\mathcal N}(0,1)\right] =\wek{0} \\
E\left[ \wek{p}_c^{(\tau)} {\mathcal N}(0,1)]\right] =\wek{0} \\
E\left[(1-c_{cov})^{\frac{t}{2}} {\mathcal N}(\wek{0},\mat{I})\right] = \wek{0}
\end{align}
therefore
\begin{align}
E\left[\wek{\delta}^{(t+1)}\right] = \wek{0}
\label{eq:expected}
\end{align}

Note that for a scalar $a$, a vector $\wek{d}$, and a standard normal variate $r \sim {\mathcal N}(0,1)$, the mean and covariance matrix of the vector $ar\wek{d}$ is defined as
\begin{align}
E[(ar\wek{d})]=E[r]\cdot E[a\wek{d}]=\wek{0}  \\
\Sigma[ar\wek{d}]=E[(ar\wek{d})(ar\wek{d})^\tran]=a^2(\wek{d})(\wek{d})^\tran 
\label{eq:covrd}
\end{align}
For a vector $\wek{d}$, a standard normal variate $r \sim {\mathcal N}(0,\mat{I})$, and a standard normal multivariate  $\wek{v} \sim {\mathcal N}(0,1)$, it holds upon their mutual independence
\begin{align}
E[r\wek{v}^\tran]=\wek{0} \\
E[(r\wek{d})\wek{v}^\tran]=\mat{0}
\label{eq:covrdv}
\end{align}
If $\wek{d}_1, \wek{d}_2$ are vectors and  $r_1,r_2 \sim {\mathcal N}(0,1)$ are independent standard normal variates then
\begin{align}
E[r_1r_2]=0 \\
E[(r_1\wek{d_1})(r_2\wek{d_2})^\tran]=\mat{0}
\label{eq:covrdrd}
\end{align}

Since the expectation vector is zero, the covariance matrix of the vector $\delta^{t+1}$ is given by:
\begin{align}
\Sigma[\wek{\delta}^{(t+1)}]=E[\wek{\delta}^{(t+1)}(\wek{\delta}^{(t+1)})^\tran]
\end{align}
Bearing in mind \eqref{eq:covrd} -- \eqref{eq:covrdrd} we get
\begin{align}
\Sigma[\wek{\delta}^{(t+1)}] = 
&\sum_{\tau=1}^{t}(1-c_{cov})^{t-\tau} \left(c_\mu \sum_{j=1}^\mu w_j \wek{d}_j^{(\tau)}(\wek{d}_j^{(\tau)})^\tran  + c_1 \wek{p}_c^{(\tau)}(\wek{p}_c^{(\tau)})^\tran \right) \nonumber \\
&+(1-c_{cov})^t \mat{I}
\end{align}
Together with Observation 1, this proves Theorem 1. {\hfill$\square$}

Theorem 1 allows for the generation of new difference vectors without requiring covariance matrix factorization. Instead, new difference vectors can be defined as a randomized weighted combination of past difference vectors. 

Note that the contribution of difference vectors, which have been generated $\tau$ iterations before the current one, diminishes by a factor of $(1 - c_{cov})^{\tau/2}$. Consequently, it is sufficient to consider only the difference vectors from the most recent $h$ populations, without incurring a significant difference in the resulting covariance matrix, compared to using all previous difference vectors. Therefore, formula \eqref{eq:th1} can be approximated as:

\begin{align}
\wek{\delta}_h^{t}\approx&\sum_{\tau=t-h+1}^{t}(1-c_{cov})^{\frac{t-\tau}{2}} \left(c_\mu^{\frac{1}{2}} \sum_{j=1}^\mu w_j^{\frac{1}{2}} \wek{d}_j^{(\tau)}  {\mathcal N}(0,1) + c_1^{\frac{1}{2}} \wek{p}_c^{(\tau)} {\mathcal N}(0,1)\right) \nonumber \\
& +(1-c_{cov})^\frac{t}{2} {\mathcal N}(\wek{0},\mat{I})
\label{eq:th1-approx}
\end{align}

The value of $h$ will be referred to as the \textit{history window size}.

Leveraging this approach, we propose the Matrix-free CMA-ES (MF-CMA-ES)  algorithm (see Figure \ref{alg:nm-cmaes}) that adapts the probability distribution to generate difference vectors using a Gaussian multivariate distribution.

\begin{figure}[h!]
\begin{algorithmic}[1]
\STATE $\wek{p}_c^{(1)} \gets \wek{0}$
\STATE $t \gets 1$
\STATE initialize$(\wek{m}^{(1)},\sigma^{(1)})$
\WHILE{!stop}
   \FOR{$i=1$ \TO $\lambda$}
      \STATE generate $ \wek{d}_i^{(t)} \sim \wek{\delta}_h^{t}$ (see \eqref{eq:th1-approx})
      \STATE $ \wek{x}_i^{(t)} = \wek{m}^{(t)} + \sigma^{(t)} \wek{d}_i^{(t)} $
   \ENDFOR
   \STATE evaluate $(X^{(t)})$
   \STATE sort $(X^{(t)})$ according to fitness
   \STATE reorder $(D^{(t)})$ according to $(X^{(t)})$   
   \STATE $\wek{\Delta}^{(t)} \gets \sum_{i=1}^\mu w_i \wek{d}_i^{(t)} $
   \STATE $\wek{m}^{(t+1)} \gets \wek{m}^{(t)} + \sigma^{(t)} \wek{\Delta}^{(t)}$
   \STATE $\wek{p}_c^{(t+1)} \gets (1-c_c)\wek{p}_c^{(t)} + \sqrt{\mu_\text{eff} c_c(2-c_c)} \cdot \wek{\Delta}^{(t)}$
   \STATE store $\left\{ \wek{d}_1^{(t)}, \ldots, \wek{d}_\mu^{(t)}, \wek{p}_c^{(t)} \right\}$ into archive
   \STATE $\sigma^{(t+1)} \gets $ update $(\sigma^{(t)} )   $
      
   \STATE $t \gets t+1$
\ENDWHILE

\end{algorithmic}
\caption{Outline of the matrix-free CMA-ES (MF-CMA-ES)}
\label{alg:nm-cmaes}
\end{figure}

MF-CMA-ES does not need matrix decomposition. To generate a population, it requires drawing $\lambda (\mu + 1)h + \lambda n$ samples from the standard normal distribution, as can be seen in formula \eqref{eq:th1-approx}. 

\section{Comparison of matrix-based and matrix-free CMA-ES without step-size adaptation}

In this section, we compare matrix-based and matrix-free covariance matrix adaptation rule, therefore, we turn off the step-size adaptation procedure, setting $\sigma=1$ in all iterations of both compared algorithms. The parameters of both algorithms are set to values recommended in \cite{Hansen2016Tutorial}, with two exceptions: the weights were equal to $1/\mu$ and the population size was $\lambda=4n$, following \cite{SMAES2017}.
For the comparison we use the quadratic fitness function \cite{experimentumCrucis}:
\begin{equation}
    q(\wek{x})=\sum_{i=1}^n 10^{6 \frac{i-1}{n-1}} x^2
    \label{eq:quadratic}
\end{equation}

\subsection{Analysis of the convergence curves}

\begin{figure}[b]
\includegraphics[width=\linewidth]{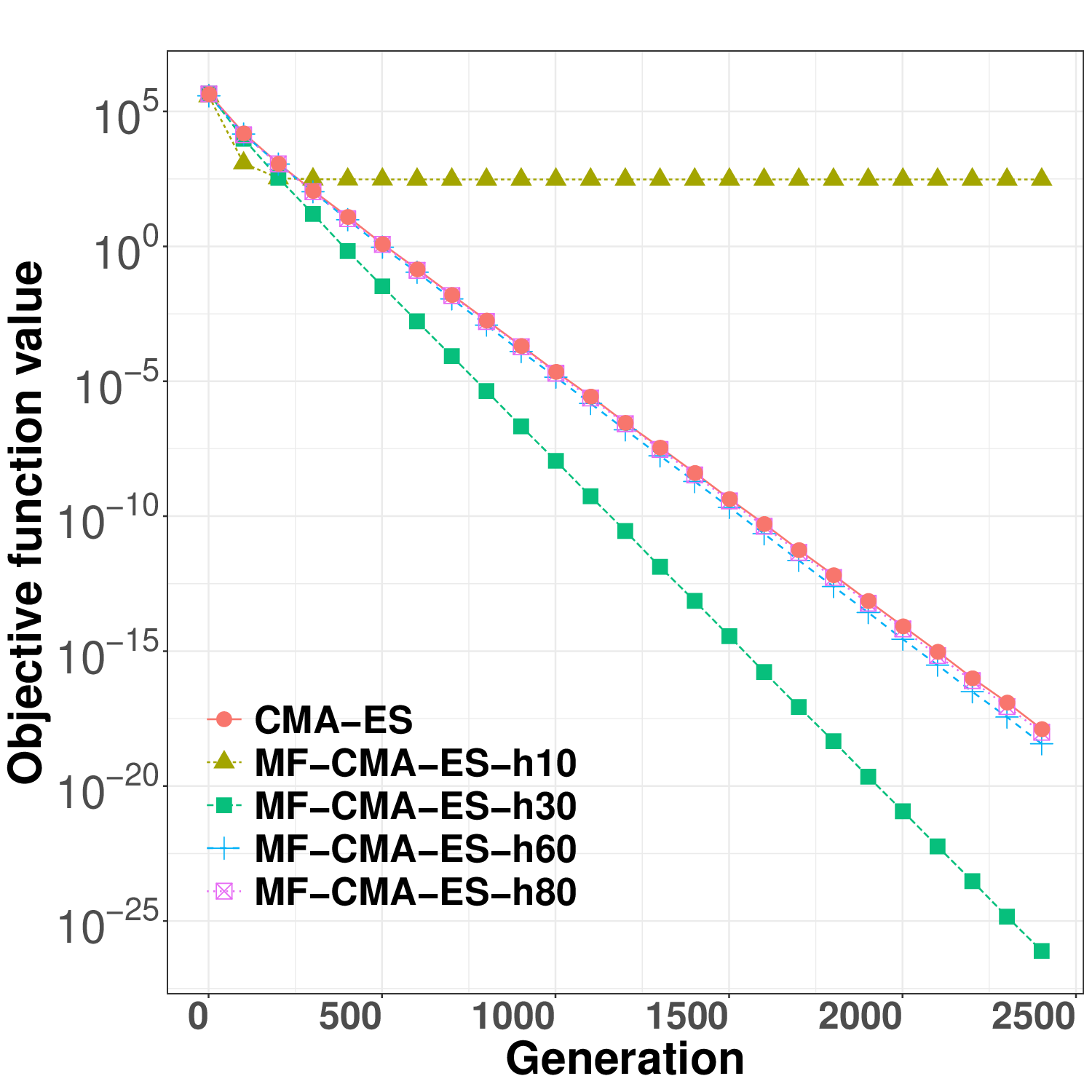}
\caption{Averaged convergence curves obtained on 30 independent runs of CMA-ES and MF-CMA-ES with different history window sizes on function \eqref{eq:quadratic}.}
\label{fig:convergence-quadratic}
\end{figure}

We report the convergence curves of both methods, assuming that both methods are allowed the same budget of fitness evaluations. The convergence curves are averaged over 30 independent runs of compared algorithms. The starting points were randomly generated with a uniform distribution from the range $[-0.2,0.8]^n$. Figure \ref{fig:convergence-quadratic} presents the convergence curves for the dimensionality $n=30$. In the case of MF-CMA-ES, we report convergence curves for the following history window sizes: $h\in \{10, 30, 60, 80\}$.

Depending on the history window size $h$, the dynamics of MF-CMA-ES becomes similar or different than of vanilla CMA-ES. When $h$ is small, the convergence curve initially goes down faster, but then it stagnates at a certain level. For sufficiently large $h$, MF-CMA-ES converges linearly. The convergence is faster than that of CMA-ES for smaller $h$. When the history window size grows, the convergence rate stabilizes at a level that is comparable to the one of CMA-ES.

\subsection{Dynamics of the covariance matrix eigenvalues}

An important feature of CMA-ES is its contour-fitting property. After a certain stabilization period of the point-generating distribution, the covariance matrix maintained by CMA-ES becomes proportional to the inverse Hessian of the fitness function.

\begin{figure}[]
{\centering
\includegraphics[width=0.75 \linewidth]{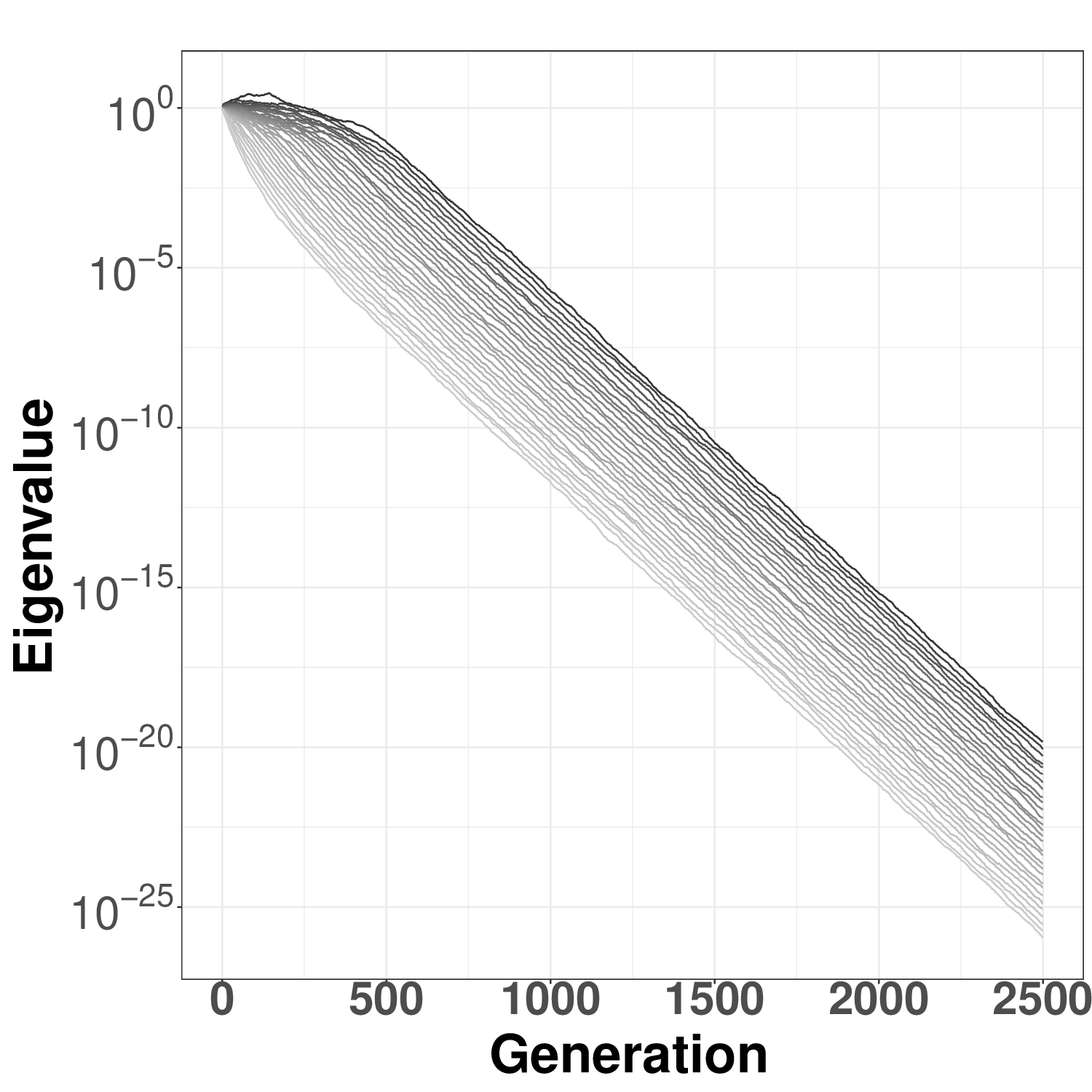}

(a)

\includegraphics[width=0.75\linewidth]{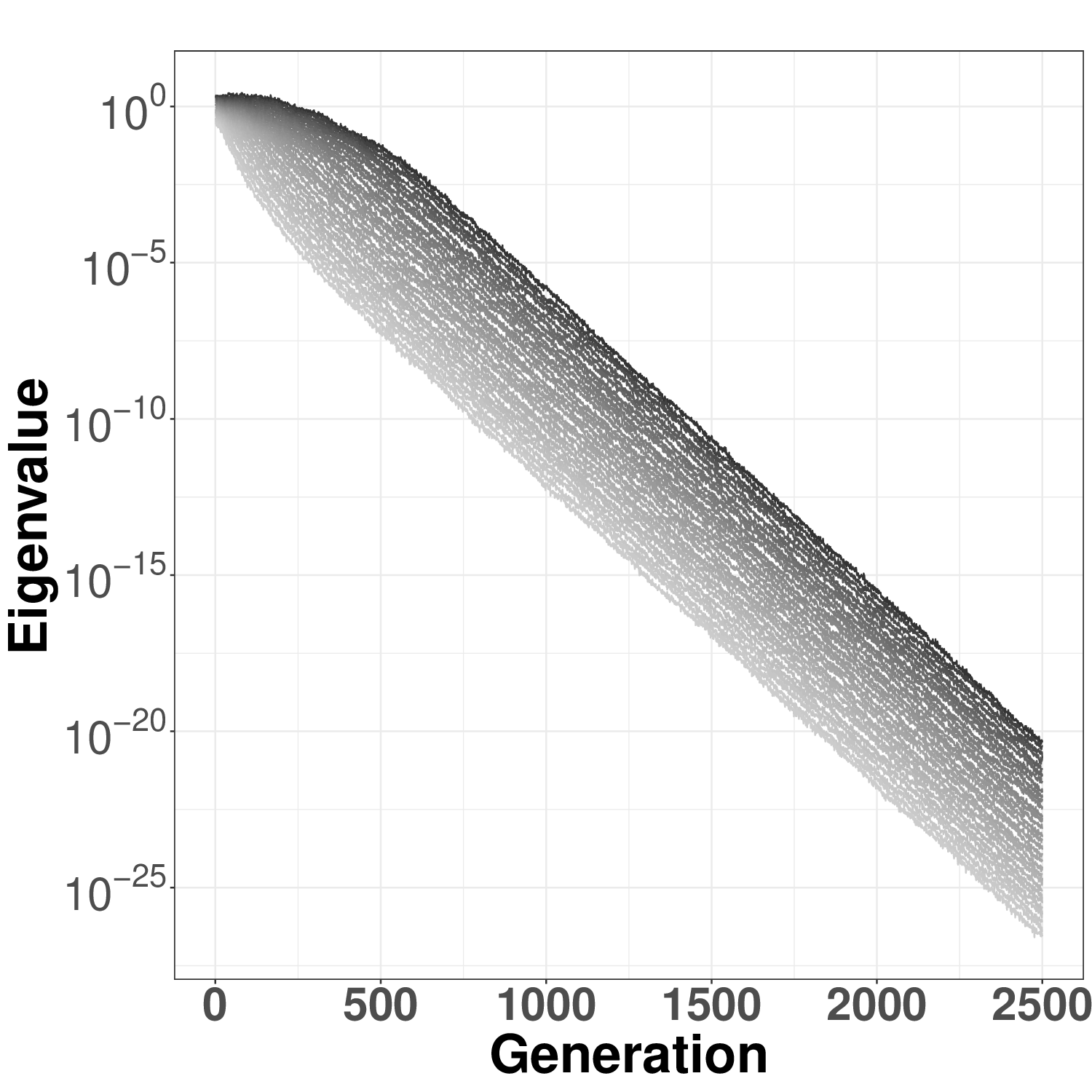}

(b)

\includegraphics[width=0.75\linewidth]{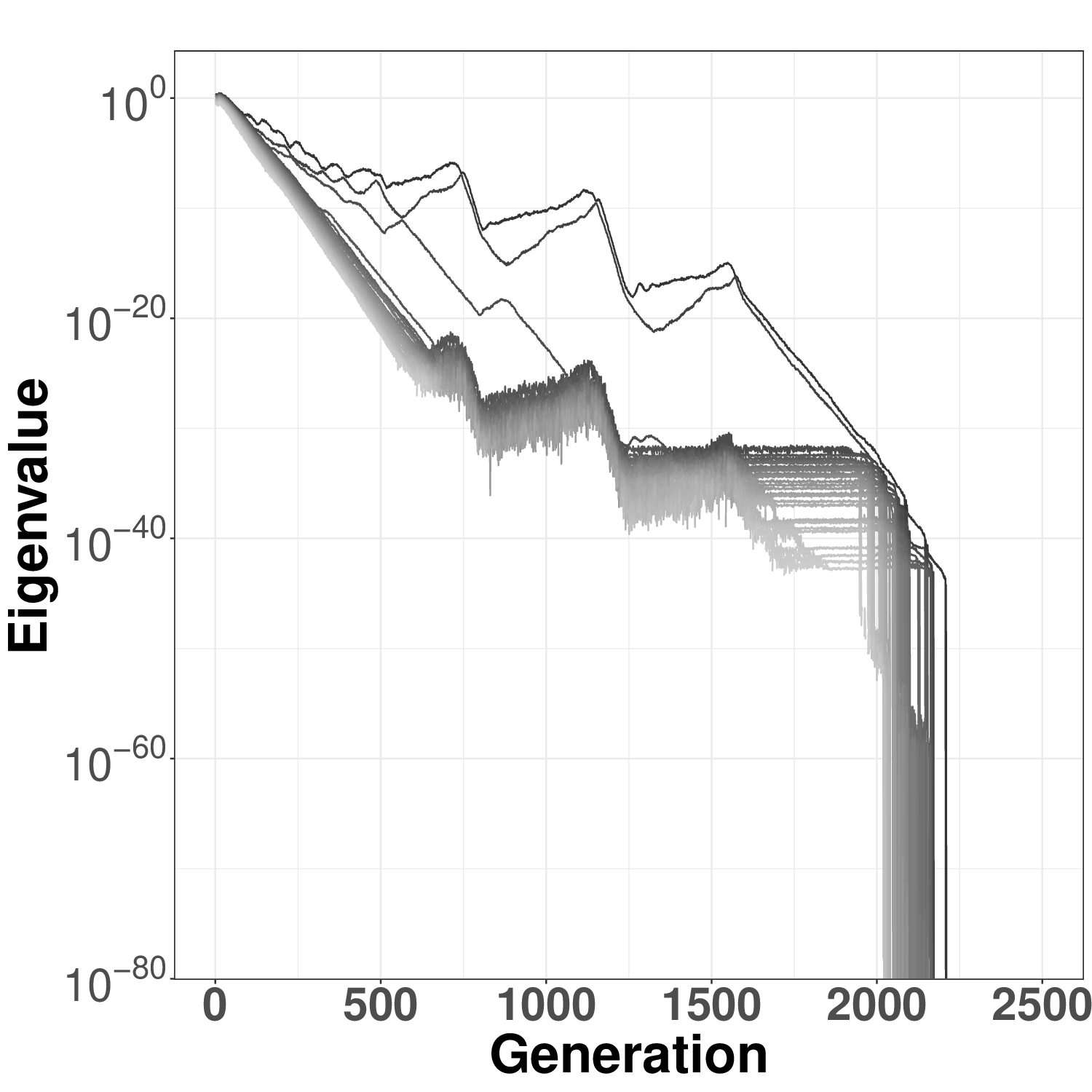}

(c)
}

\caption{Dynamics of eigenvalues for the vanilla CMA-ES (a) and for the MF-CMA-ES with the window size $h=60$ (b) and $h=10$ (c)}
\label{fig:eigenplots}
\end{figure}

Following the methodology from \cite{experimentumCrucis}, in each iteration of both compared algorithms, we computed the eigenvalues of the empirical covariance matrix of the generated difference vectors (MF-CMA-ES) and of the covariance matrix $C^{(t)}$ (vanilla CMA-ES). Figure \ref{fig:eigenplots} shows the dynamics of the eigenvalues from a single run of both methods. For MF-CMA-ES, we demonstrate two different history window sizes: $h=60$ and
$h=10$. The dimension was set to $n=30$.

The eigenvalues dynamics for the vanilla CMA-ES and the MF-CMA-ES with $h=60$ are similar, indicating that both methods generate comparable point distributions. 

For $h=10$, the algorithm fails to capture the desired eigenvalues dynamics. The high condition number of the covariance matrix suggests that the population is confined to a low-dimensional subspace, which becomes a fundamental obstacle to reaching the optimum and explains the poor convergence observed for $h=10$.

\section{Matrix-free CMA-ES with step-size adaptation}

For CMA-ES it has been observed that coupling the covariance matrix adaptation procedure with an additional step-size adaptation mechanism can significantly improve the global optimization efficiency of the algorithm. A Cumulative Step Adaptation (CSA) rule \cite{hansen2006cma} is an efficient and widely accepted method to address this problem. Unfortunately, the matrix-free CMA-ES cannot be coupled with CSA, as CSA requires the eigendecomposition of the covariance matrix to compute its inverse square root. Therefore, we decided to use the Previous Population Midpoint Fitness (PPMF) method \cite{Warchulski2021} to control the step size in the matrix-free CMA-ES.

\subsection{Cumulative Step Adaptation (CSA)}

The CSA procedure (Figure \ref{alg:CSA}) analyzes the set of $\wek{z}_i^{(t)}$ vectors that are ordered in the same sequence as their corresponding points $\wek{x}_i^{(t)}$ that are sorted with respect to their fitness. The procedure tracks the value of the vector $\wek{p}_\sigma^{(t)}$, which is changed in each iteration by accumulating the mean value of $\mu$ vectors $\wek{z}_i^{(t)}$ corresponding to the $\mu$ best individuals generated in the current generation. If the fitness function is flat, a random walk of the population midpoint will be observed, and the norm of $\wek{p}_\sigma$ will be $\chi$ distributed with $n$ degrees of freedom. Then the expectation of the logarithm of step-size change will be zero. When the midpoint $\wek{m}^t$ is located in a significant distance from the local optimum, consecutive shifts of the midpoint will be positively correlated, which will increase $||\wek{p}_\sigma||$ and the step size will also increase. Contrarily, if the midpoint $\wek{m}^t$ is located close to the minimum, then, on the average, shorter difference vectors will be correlated with better-fit points. In effect, $||\wek{p}_\sigma||$ and the step size will be decreased.

\begin{figure}[h!]
\begin{algorithmic}[1]
   
   \STATE $\wek{\Delta}_\sigma^{(t)} \gets \sum_{i=1}^\mu w_i \wek{z}_i^{(t)} $

   \STATE $\wek{p}_\sigma^{(t+1)} \gets (1-c_\sigma)\wek{p}_\sigma^{(t)} + \sqrt{\mu c_\sigma(2-c_\sigma)} \cdot \wek{\Delta}_\sigma^{(t)}$
   \STATE $\sigma^{(t+1)} \gets \sigma^{(t)} \exp\left(\frac{c_\sigma}{d_\sigma} \left(\frac{\| \wek{p}_\sigma^{(t+1)} \|}{E\|N(\wek{0},\mat{I})\|}-1\right)\right) $

\end{algorithmic}
\caption{Outline of the Cumulative Step Adaptation (CSA)}
\label{alg:CSA}
\end{figure}

\subsection{Previous Point Midpoint Fitness (PPMF)}

PPMF is an alternative step-size adaptation method inspired by the one-fifth rule. Pseudocode of the method is given in Figure \ref{alg:PPMF}. In each iteration, fitness of the previous population midpoint is evaluated by the fitness function denoted as $q\colon \mathbb{R}^{n} \rightarrow \mathbb{R}$. The fitness of the points of the current population is compared to the fitness of the previous population midpoint. If the number of points superior to the midpoint is a small percentage of the whole population then the step-size is reduced. The step-size is increased when the proportion of superior points exceeds a threshold value $\theta$, a PPMF hyperparameter. According to the results reported in \cite{Warchulski2021}, the method achieves good efficiency with $\theta=0.2$. We used this value as is, without any further hyperparameter tuning.

\begin{figure}[h]
\begin{algorithmic}[1]
    \STATE $\overline{\wek{m}}^{(t-1)} \gets \frac{1}{\lambda}\sum_{i=1}^\lambda \wek{X}_i^{(t-1)}$
   \STATE evaluate $(\overline{\wek{m}}^{t-1})$
   \STATE $p^{(t)}_{s} \gets \left|\{i: q(\wek{x}_{i}^{t}) < q(\overline{\wek{m}}^{(t-1)})\}\right|/\lambda$
   \STATE $\sigma^{(t+1)} \gets \sigma^{(t)} \exp\left(\frac{1}{d_\sigma} \cdot \frac{p^{(t)}_s-\theta}{1-\theta}\right) $
\end{algorithmic}
\caption{Outline of the Previous Population Midpoint Fitness (PPMF) step-size adaptation}
\label{alg:PPMF}
\end{figure}

\subsection{Benchmarking matrix-free and matrix-based CMA-ES with step-size adaptation}

We evaluated the efficiency of the newly introduced MF-CMA-ES algorithm using the CEC'2017 benchmark suite \cite{suganthan2017problem}. In line with the CEC'2017 benchmarking guidelines, 51 independent runs were conducted for each optimization problem. For each run, the fitness evaluation budget was set to MaxFEs = $10000n$.

In the experiments, MF-CMA-ES was coupled with the PPMF step size adaptation method. For comparison, we used two versions of vanilla CMA-ES, coupled with either CSA or PPMF step-size control. In all optimization problems from the CEC'2017 suite, all common parameters of considered algorithms were set to identical values.

The history window size depended on the dimension number according to the formula $h = 20 + 1.4n$. We derived this heuristic by determining, for each value of $n$, the smallest values of $h$ that resulted in the convergence rate of MF-CMA-ES similar to that of vanilla CMA-ES on the function \eqref{eq:quadratic}, assuming constant $\sigma$ for both methods.

Tables \ref{tab:CEC17-D10} -- \ref{tab:CEC17-D50} present the summary of results obtained for each problem from the CEC'2017 suite. The results show the error, i.e. the difference between the solution objective function value yielded by the algorithm and the objective function value at the global optimum. For each optimization problem, the mean and standard deviation of 51 smallest error values achieved in each run after exceeding the fitness evaluation budget are reported.

 To compare the optimization algorithms, we used statistical tests recommended by \cite{DERRAC20113}. First, we applied the Quade test to determine whether a significant performance difference existed between at least one pair of algorithms. The p-values of the null hypothesis were equal $0.0003$ for $n=10$, $0.08$ for $n=30$, and $0.008$ for $n=50$. Hence, for $n=30$, the null hypothesis cannot be rejected, indicating no significant difference between the algorithms.

The Quade test provides insights into the ranking of the compared methods. Based on this ranking, in $n=10$ dimensions, MF-CMA-ES outperformed  both versions of matrix-based CMA-ES. However, for $n=50$, CMA-ES-CSA was superior to MF-CMA-ES, while MF-CMA-ES outperformed CMA-ES-PPMF.

Next, we compared MF-CMA-ES with best-performing CMA-ES variant and analyzed the Holm-corrected p-values. The results were produced using the software provided by the authors of \cite{GARCIA20102044}. The Holm-corrected p-values were 0.034 for $n=10$ and 0.013 for $n=50$. Assuming a significance level of $\alpha = 0.05$, the null hypothesis was rejected for $n=10$, indicating the superiority of MF-CMA-ES, and for $n=50$ indicating the superiority of CMA-ES-CSA.

\begin{figure}[b]
\centering
\includegraphics[width=0.95 \linewidth]{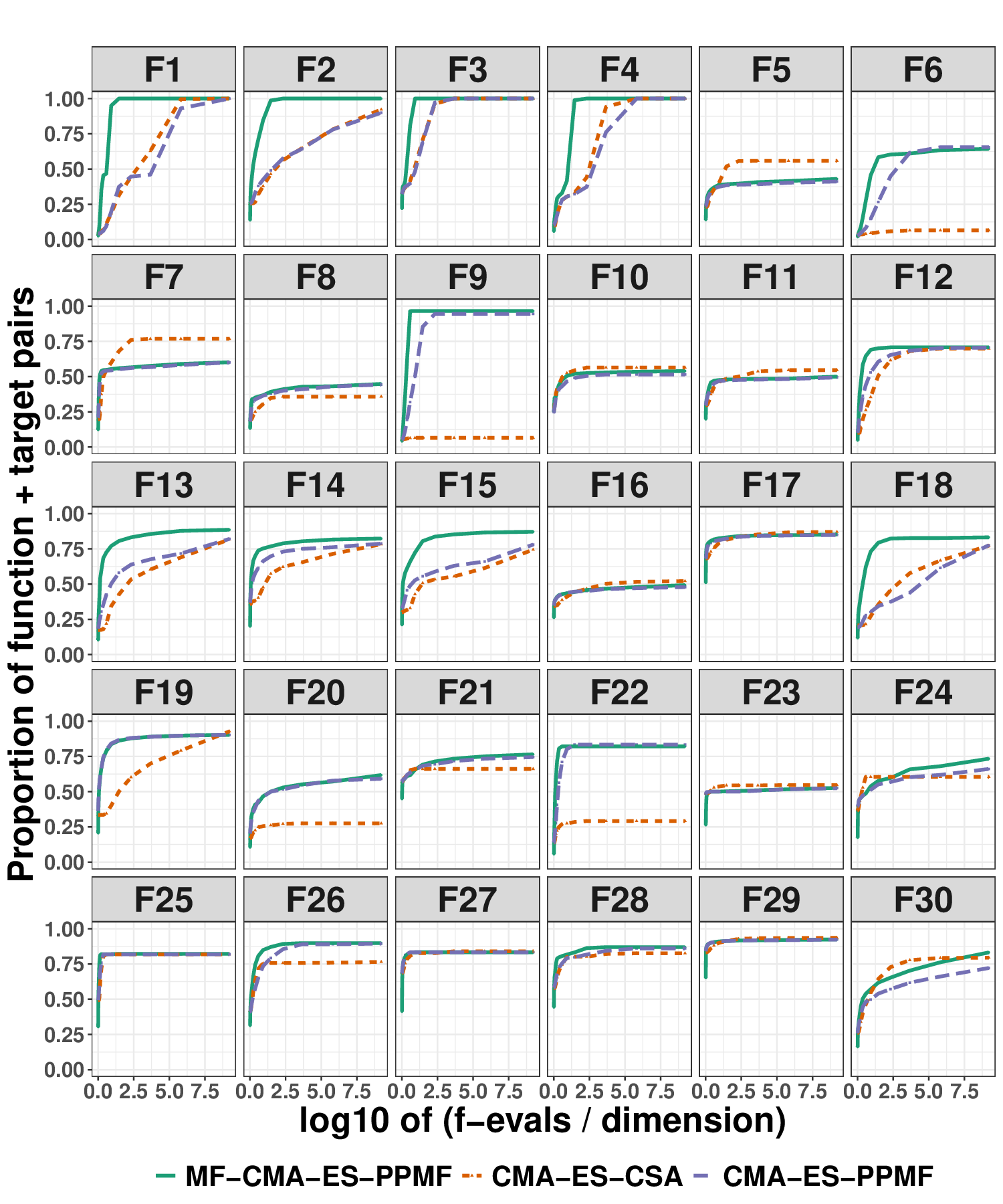}

\caption{ECDF curves obtained by MF-CMA-ES (with PPMF) and two versions of CMA-ES (with CSA and PPMF) for CEC'2017 optimization problems in 10 dimensions}
\label{fig:CEC17-d10}
\end{figure}

\begin{figure}
\centering
\includegraphics[width=0.95\linewidth]{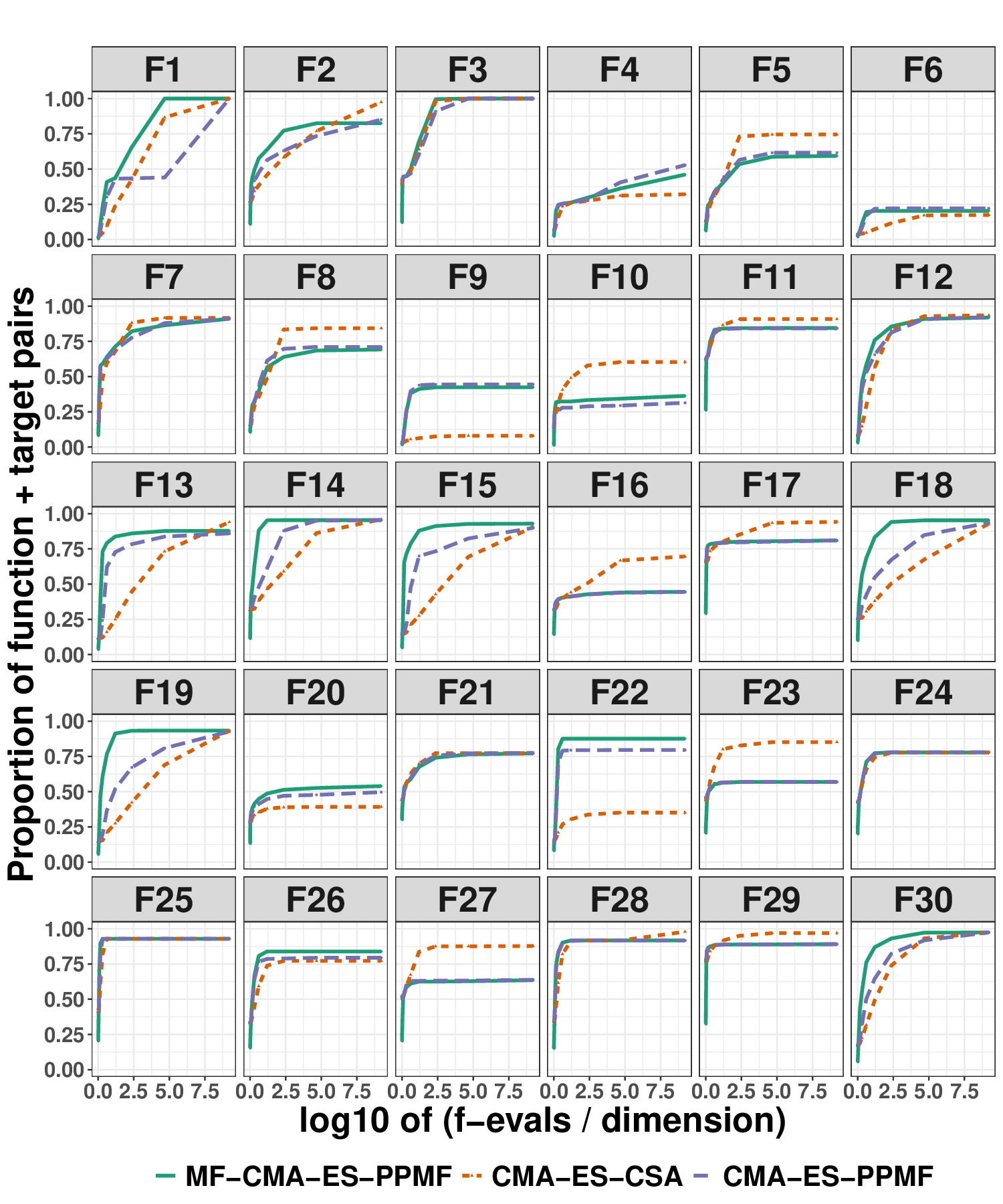}

(a) 

\includegraphics[width=0.95\linewidth]{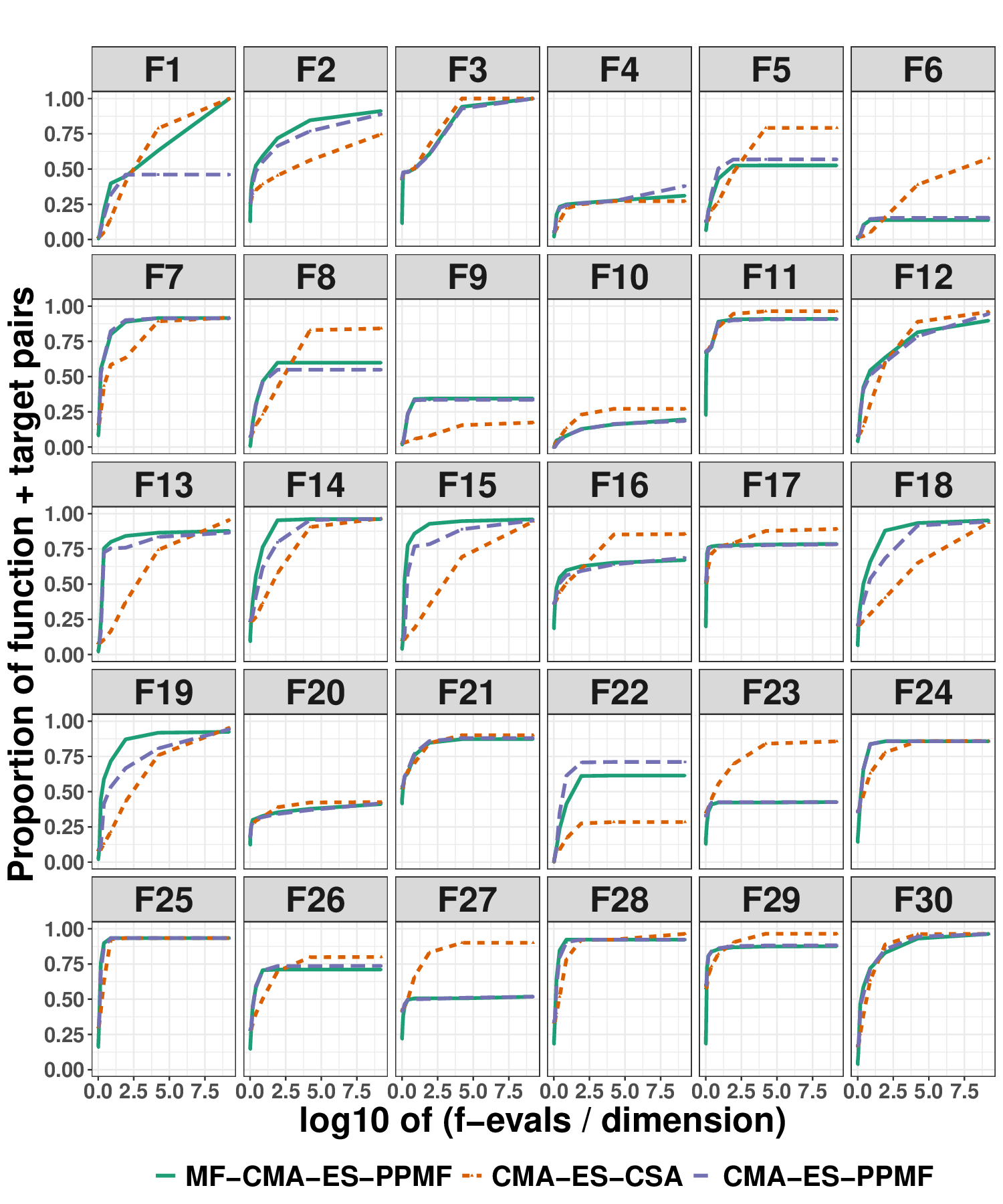}

(b)
\caption{ECDF curves obtained by MF-CMA-ES (with PPMF) and two versions of CMA-ES (with CSA and PPMF) for CEC'2017  optimization problems in 30 dimensions (a) and 50 dimensions (b).}
\label{fig:CEC17-d30-d50}
\end{figure}

To facilitate a more detailed comparison between algorithms, we analyze their dynamics using Empirical Cumulative Distribution Functions (ECDFs) as described in \cite{cocoPerformance}.
For each problem and dimension, we establish a logarithmic scale of target precision values spanning the range of best and worst results achieved by any algorithm. This scale is defined with a ratio of $10^{0.2}$ between neighboring values.
Then, for each method and at every percentage of the maximum function evaluations (MaxFEs), we compute the average percentage of fitness levels achieved across all independent runs.
This yields a non-decreasing ECDF curve for each method on each problem, providing a measure of its efficiency. These curves are presented in Figures \ref{fig:CEC17-d10} and \ref{fig:CEC17-d30-d50}.

From the ECDF curves, we observe that the dynamics of all compared algorithms were similar in many cases. In some cases (e.g. F1 in $n=10, 30$) the final results were of a comparable quality for all methods, but MF-CMA-ES converged faster. Only in one case (F22, $n=50$) the CMA-ES-PPMF significantly outperformed MF-CMA-ES. We conclude that MF-CMA-ES should be preferred over CMA-ES-PPMF, especially when the budget for fitness evaluations is smaller than that assumed in CEC'2017.

\input{tables/cec-2017-d-10-problem-table}

\section{Concluding remarks}

We introduced MF-CMA-ES, a matrix-free version of CMA-ES that eliminates the use of covariance matrix for generating new individuals. This work provides a fresh perspective on CMA-ES by demonstrating that the covariance matrix adaptation process can be performed implicitly. MF-CMA-ES generates new individuals through a weighted combination of points from the archive of previous generations. The proposed method preserves the distributional properties of the original CMA-ES.

\input{tables/cec-2017-d-30-problem-table}
\input{tables/cec-2017-d-50-problem-table}

The differences between convergence dynamics of MF-CMA-ES and CMA-ES for the quadratic fitness function were very small when the step-size adaptation was turned off. It is likely that these subtle differences are amplified when the step-size adaptation is enabled, which may lead to discrepancies in the quality of results between matrix-free and matrix-based CMA-ES during the CEC'2017 benchmarking procedure. Overall, MF-CMA-ES typically performed at least as well as CMA-ES-PPMF.

The formula \eqref{eq:th1-approx} implies that the influence of the archive vectors decays exponentially over time. We plan to explore alternative formulations to express this influence. Specifically, a linear weight decay scheme will be considered, which could allow for smaller history window sizes, ultimately reducing both time and memory complexity.

We also plan to improve the code efficiency of MF-CMA-ES to fully leverage its vector-based form. 

\section*{Implementation note}

All experiments presented in this paper were conducted using the R programming language. 
We used the {\tt cma-es} package \cite{cmaes-package} to simulate the matrix-based CMA-ES. The source code of MF-CMA-ES and PPMF was developed by modifying the {\tt cma-es} package. Unless otherwise noted, we used the default parameter values from the {\tt cma-es} package for all tested matrix-based and matrix-free CMA-ES versions. The source code to reproduce all the results is available at \href{https://github.com/AdamStelmaszczyk/mf-cma-es}{https://github.com/AdamStelmaszczyk/mf-cma-es}. Benchmarking results provided in section 5.3 were obtained using the CEC'2017 implementation in R \cite{cecs-package}.

\clearpage

\bibliography{DESTEC}

\end{document}

%% file: tables/cec-2017-d-10-problem-table.tex
\begin{table}[t!]
\centering
\resizebox{0.5\textwidth}{!}{\begin{tabular}{c|c|c|c|c|c|c}
\toprule
\multicolumn{1}{c|}{\textbf{F.}} & \multicolumn{2}{c|}{\textbf{MF-CMA-ES}} & \multicolumn{2}{c|}{\textbf{CMA-ES-CSA}} & \multicolumn{2}{c}{\textbf{CMA-ES-PPMF}}\\
\midrule
{} & Mean & Std & Mean & Std & Mean & Std\\
\hline

1 & \textbf{0.0e+00} & \textbf{0.0e+00} & \textbf{0.0e+00} & \textbf{0.0e+00} & \textbf{0.0e+00} & \textbf{0.0e+00}\\
\hline
2 & \textbf{0.0e+00} & \textbf{0.0e+00} & 2.1e-06 & 1.5e-06 & 8.8e+09 & 6.3e+10\\
\hline
3 & \textbf{0.0e+00} & \textbf{0.0e+00} & \textbf{0.0e+00} & \textbf{0.0e+00} & \textbf{0.0e+00} & \textbf{0.0e+00}\\
\hline
4 & \textbf{0.0e+00} & \textbf{0.0e+00} & \textbf{0.0e+00} & \textbf{0.0e+00} & \textbf{0.0e+00} & \textbf{0.0e+00}\\
\hline
5 & \textbf{4.8e+01} & \textbf{6.4e+00} & 7.1e+01 & 7.6e+01 & 5.5e+01 & 2.6e+01\\
\hline
6 & 2.1e+00 & 1.4e+01 & 7.5e+01 & 2.7e+01 & \textbf{1.8e+00} & \textbf{1.2e+01}\\
\hline
7 & 4.5e+01 & 4.7e+00 & \textbf{2.8e+01} & \textbf{7.5e+01} & 4.7e+01 & 5.3e+00\\
\hline
8 & \textbf{3.6e+01} & \textbf{8.7e+00} & 8.6e+01 & 5.1e+01 & \textbf{3.6e+01} & \textbf{8.3e+00}\\
\hline
9 & 1.2e-02 & 6.6e-02 & 3.4e+03 & 1.3e+03 & \textbf{8.8e-03} & \textbf{2.7e-02}\\
\hline
10 & \textbf{1.5e+03} & \textbf{1.8e+02} & 1.6e+03 & 3.7e+02 & 1.7e+03 & 2.1e+02\\
\hline
11 & 9.3e+00 & 6.7e+00 & \textbf{6.1e+00} & \textbf{7.5e+00} & 1.0e+01 & 9.1e+00\\
\hline
12 & \textbf{4.3e+02} & \textbf{2.2e+02} & 4.4e+02 & 1.9e+02 & 4.8e+02 & 2.4e+02\\
\hline
13 & \textbf{1.3e+01} & \textbf{3.4e+00} & 8.0e+01 & 6.6e+01 & 7.9e+01 & 8.1e+01\\
\hline
14 & \textbf{2.5e+01} & \textbf{8.7e+00} & 5.6e+01 & 1.7e+01 & 5.0e+01 & 1.0e+01\\
\hline
15 & \textbf{6.6e+00} & \textbf{1.3e+01} & 1.0e+02 & 4.9e+01 & 4.2e+01 & 2.1e+01\\
\hline
16 & \textbf{2.3e+02} & \textbf{7.2e+01} & 3.8e+02 & 2.7e+02 & 2.7e+02 & 8.1e+01\\
\hline
17 & \textbf{1.0e+02} & \textbf{3.9e+01} & 1.3e+02 & 1.5e+02 & \textbf{1.0e+02} & \textbf{1.5e+01}\\
\hline
18 & \textbf{4.2e+01} & \textbf{3.5e+01} & 1.3e+02 & 7.9e+01 & 1.2e+02 & 5.1e+01\\
\hline
19 & 5.1e+01 & 1.5e+01 & \textbf{3.2e+01} & \textbf{2.0e+01} & 4.9e+01 & 1.6e+01\\
\hline
20 & \textbf{1.7e+02} & \textbf{4.6e+01} & 4.3e+02 & 1.3e+02 & 1.8e+02 & 4.9e+01\\
\hline
21 & \textbf{1.7e+02} & \textbf{3.6e+01} & 2.5e+02 & 8.1e+01 & 1.8e+02 & 2.6e+01\\
\hline
22 & 2.3e+02 & 5.0e+02 & 1.4e+03 & 7.0e+02 & \textbf{2.0e+02} & \textbf{4.2e+02}\\
\hline
23 & 4.3e+02 & 2.0e+02 & 4.6e+02 & 1.3e+02 & \textbf{4.2e+02} & \textbf{5.4e+01}\\
\hline
24 & \textbf{1.8e+02} & \textbf{8.1e+01} & 3.1e+02 & 8.6e+01 & 2.4e+02 & 8.5e+01\\
\hline
25 & \textbf{4.3e+02} & \textbf{3.8e+01} & 4.4e+02 & 1.2e+01 & \textbf{4.3e+02} & \textbf{2.0e+01}\\
\hline
26 & \textbf{2.9e+02} & \textbf{3.4e+01} & 8.2e+02 & 8.4e+02 & \textbf{2.9e+02} & \textbf{3.8e+01}\\
\hline
27 & 4.7e+02 & 1.6e+01 & \textbf{4.2e+02} & \textbf{7.6e+01} & 4.8e+02 & 1.7e+01\\
\hline
28 & \textbf{4.6e+02} & \textbf{1.4e+02} & 5.4e+02 & 1.3e+02 & 4.7e+02 & 1.4e+02\\
\hline
29 & 4.0e+02 & 3.4e+01 & \textbf{3.9e+02} & \textbf{1.6e+02} & 4.1e+02 & 3.7e+01\\
\hline
30 & 4.9e+05 & 8.8e+05 & \textbf{4.1e+05} & \textbf{5.2e+05} & 1.2e+06 & 1.4e+06\\

\bottomrule
\end{tabular}
}
\caption{Statistics of results obtained by MF-CMA-ES (with PPMF) and two versions of CMA-ES (with CSA and PPMF) for CEC'2017 optimization problems in 10 dimensions after spending the budget of $10000n$ fitness evaluations.}
\label{tab:CEC17-D10}
\end{table}

%% file: tables/cec-2017-d-30-problem-table.tex
\begin{table}[t!]
\centering
\resizebox{0.5\textwidth}{!}{\begin{tabular}{c|c|c|c|c|c|c}
\toprule
\multicolumn{1}{c|}{\textbf{F.}} & \multicolumn{2}{c|}{\textbf{MF-CMA-ES}} & \multicolumn{2}{c|}{\textbf{CMA-ES-CSA}} & \multicolumn{2}{c}{\textbf{CMA-ES-PPMF}}\\
\midrule
{} & Mean & Std & Mean & Std & Mean & Std\\
\hline
1 & \textbf{0.0e+00} & \textbf{0.0e+00} & \textbf{0.0e+00} & \textbf{0.0e+00} & \textbf{0.0e+00} & \textbf{0.0e+00}\\
\hline
2 & 4.1e+46 & 2.9e+47 & \textbf{8.0e-07} & \textbf{6.0e-07} & 3.1e+44 & 1.8e+45\\
\hline
3 & \textbf{0.0e+00} & \textbf{0.0e+00} & \textbf{0.0e+00} & \textbf{0.0e+00} & \textbf{0.0e+00} & \textbf{0.0e+00}\\
\hline
4 & 1.8e+01 & 2.7e+01 & 5.5e+01 & 1.4e+01 & \textbf{1.4e+01} & \textbf{2.4e+01}\\
\hline
5 & 5.5e+01 & 1.3e+02 & \textbf{3.1e+01} & \textbf{1.1e+02} & 3.3e+01 & 7.5e+01\\
\hline
6 & 1.1e+01 & 2.7e+01 & 6.2e+01 & 3.3e+01 & \textbf{7.0e+00} & \textbf{2.2e+01}\\
\hline
7 & 4.2e+01 & 2.2e+01 & \textbf{3.6e+01} & \textbf{1.7e+00} & 4.2e+01 & 2.4e+01\\
\hline
8 & 6.1e+01 & 1.4e+02 & \textbf{2.6e+01} & \textbf{8.7e+01} & 4.5e+01 & 1.1e+02\\
\hline
9 & 1.0e+00 & 1.1e+00 & 1.3e+04 & 3.6e+03 & \textbf{9.8e-01} & \textbf{1.7e+00}\\
\hline
10 & 6.4e+03 & 1.5e+03 & \textbf{4.0e+03} & \textbf{6.7e+02} & 6.8e+03 & 1.5e+03\\
\hline
11 & 1.3e+02 & 4.7e+01 & \textbf{3.6e+01} & \textbf{2.9e+01} & 1.4e+02 & 4.0e+01\\
\hline
12 & 1.5e+03 & 5.1e+02 & \textbf{1.1e+03} & \textbf{4.3e+02} & 1.4e+03 & 4.0e+02\\
\hline
13 & 1.6e+03 & 7.4e+02 & \textbf{3.8e+02} & \textbf{2.0e+02} & 2.1e+03 & 7.8e+02\\
\hline
14 & 1.4e+02 & 4.7e+01 & \textbf{1.2e+02} & \textbf{3.5e+01} & 1.3e+02 & 3.3e+01\\
\hline
15 & \textbf{2.1e+02} & \textbf{9.4e+01} & 3.4e+02 & 8.9e+01 & 3.8e+02 & 1.6e+02\\
\hline
16 & 1.7e+03 & 3.0e+02 & \textbf{2.1e+02} & \textbf{1.8e+02} & 1.8e+03 & 2.9e+02\\
\hline
17 & 5.8e+02 & 1.0e+02 & \textbf{8.1e+01} & \textbf{1.3e+02} & 5.7e+02 & 1.4e+02\\
\hline
18 & \textbf{1.5e+02} & \textbf{6.3e+01} & 2.5e+02 & 8.5e+01 & 2.3e+02 & 8.8e+01\\
\hline
19 & \textbf{1.1e+02} & \textbf{3.7e+01} & 1.2e+02 & 4.7e+01 & 1.3e+02 & 4.4e+01\\
\hline
20 & \textbf{8.1e+02} & \textbf{2.1e+02} & 1.2e+03 & 2.9e+02 & 8.9e+02 & 2.2e+02\\
\hline
21 & 2.4e+02 & 7.6e+01 & \textbf{2.2e+02} & \textbf{4.1e+01} & 2.3e+02 & 1.2e+01\\
\hline
22 & \textbf{4.2e+02} & \textbf{1.3e+03} & 4.1e+03 & 1.7e+03 & 1.1e+03 & 2.3e+03\\
\hline
23 & 1.1e+03 & 1.0e+02 & \textbf{3.8e+02} & \textbf{1.6e+02} & 1.1e+03 & 1.1e+02\\
\hline
24 & 4.4e+02 & 1.5e+01 & \textbf{4.2e+02} & \textbf{6.3e+00} & 4.3e+02 & 3.7e+01\\
\hline
25 & \textbf{3.9e+02} & \textbf{7.8e+00} & \textbf{3.9e+02} & \textbf{1.2e-02} & \textbf{3.9e+02} & \textbf{1.5e+01}\\
\hline
26 & \textbf{6.2e+02} & \textbf{5.2e+02} & 7.2e+02 & 2.9e+02 & 9.1e+02 & 8.6e+02\\
\hline
27 & 1.3e+03 & 9.4e+01 & \textbf{5.1e+02} & \textbf{9.4e+00} & 1.4e+03 & 1.0e+02\\
\hline
28 & \textbf{3.2e+02} & \textbf{4.1e+01} & 3.3e+02 & 5.0e+01 & \textbf{3.2e+02} & \textbf{4.1e+01}\\
\hline
29 & 1.7e+03 & 2.6e+02 & \textbf{4.5e+02} & \textbf{6.5e+01} & 1.7e+03 & 2.3e+02\\
\hline
30 & \textbf{2.1e+03} & \textbf{9.6e+01} & 2.2e+03 & 1.5e+02 & 2.5e+03 & 2.2e+02\\
\bottomrule
\end{tabular}
}
\caption{Statistics of results obtained by MF-CMA-ES (with PPMF) and two versions of CMA-ES (with CSA and PPMF) for CEC'2017 optimization problems in 30 dimensions after spending the budget of $10000n$ fitness evaluations.}
\label{tab:CEC17-D30}
\end{table}

%% file: tables/cec-2017-d-50-problem-table.tex
\begin{table}[t!]
\centering
\resizebox{0.5\textwidth}{!}{\begin{tabular}{c|c|c|c|c|c|c}
\toprule
\multicolumn{1}{c|}{\textbf{F.}} & \multicolumn{2}{c|}{\textbf{MF-CMA-ES}} & \multicolumn{2}{c|}{\textbf{CMA-ES-CSA}} & \multicolumn{2}{c}{\textbf{CMA-ES-PPMF}}\\
\midrule
{} & Mean & Std & Mean & Std & Mean & Std\\
\hline
1 & \textbf{0.0e+00} & \textbf{0.0e+00} & \textbf{0.0e+00} & \textbf{0.0e+00} & 2.0e+03 & 2.9e+03\\
\hline
2 & \textbf{3.4e+75} & \textbf{2.4e+76} & 7.3e+79 & 5.1e+80 & 4.8e+81 & 3.5e+82\\
\hline
3 & \textbf{0.0e+00} & \textbf{0.0e+00} & \textbf{0.0e+00} & \textbf{0.0e+00} & \textbf{0.0e+00} & \textbf{0.0e+00}\\
\hline
4 & 6.6e+01 & 4.6e+01 & 9.5e+01 & 4.0e+01 & \textbf{5.2e+01} & \textbf{4.6e+01}\\
\hline
5 & 2.2e+02 & 4.0e+02 & \textbf{9.9e+00} & \textbf{2.7e+00} & 8.8e+01 & 1.7e+02\\
\hline
6 & 2.2e+01 & 3.6e+01 & 3.1e+01 & 3.9e+01 & \textbf{1.1e+01} & \textbf{2.4e+01}\\
\hline
7 & 6.5e+01 & 8.3e+00 & \textbf{6.0e+01} & \textbf{1.6e+00} & 6.5e+01 & 8.0e+00\\
\hline
8 & 8.8e+01 & 1.8e+02 & \textbf{1.0e+01} & \textbf{2.3e+00} & 1.4e+02 & 2.8e+02\\
\hline
9 & \textbf{1.2e+01} & \textbf{1.9e+01} & 2.6e+04 & 1.1e+04 & 1.3e+01 & 1.7e+01\\
\hline
10 & 6.8e+03 & 1.9e+03 & \textbf{5.1e+03} & \textbf{1.8e+03} & 7.1e+03 & 2.1e+03\\
\hline
11 & 1.8e+02 & 4.8e+01 & \textbf{4.5e+01} & \textbf{1.7e+01} & 1.9e+02 & 5.2e+01\\
\hline
12 & 1.2e+04 & 2.7e+04 & \textbf{2.2e+03} & \textbf{5.5e+02} & 3.0e+03 & 6.5e+02\\
\hline
13 & 3.0e+03 & 1.6e+03 & \textbf{6.1e+02} & \textbf{2.2e+02} & 3.9e+03 & 8.7e+02\\
\hline
14 & 2.1e+02 & 4.5e+01 & \textbf{1.9e+02} & \textbf{3.6e+01} & 2.1e+02 & 4.8e+01\\
\hline
15 & \textbf{4.9e+02} & \textbf{2.5e+02} & 6.4e+02 & 1.1e+02 & 5.8e+02 & 1.6e+02\\
\hline
16 & 1.2e+03 & 5.0e+02 & \textbf{3.2e+02} & \textbf{1.8e+02} & 1.1e+03 & 4.4e+02\\
\hline
17 & 1.6e+03 & 4.1e+02 & \textbf{2.4e+02} & \textbf{1.1e+02} & 1.7e+03 & 4.0e+02\\
\hline
18 & \textbf{2.8e+02} & \textbf{9.7e+01} & 3.7e+02 & 9.1e+01 & 3.3e+02 & 1.0e+02\\
\hline
19 & 2.8e+03 & 1.4e+04 & \textbf{9.6e+01} & \textbf{2.4e+01} & 1.2e+02 & 3.3e+01\\
\hline
20 & 1.8e+03 & 4.6e+02 & \textbf{1.7e+03} & \textbf{7.1e+02} & \textbf{1.7e+03} & \textbf{3.9e+02}\\
\hline
21 & 3.3e+02 & 3.1e+02 & \textbf{2.1e+02} & \textbf{3.2e+00} & 3.0e+02 & 2.2e+02\\
\hline
22 & 3.5e+03 & 4.3e+03 & 5.6e+03 & 2.3e+03 & \textbf{2.3e+03} & \textbf{3.7e+03}\\
\hline
23 & 2.0e+03 & 2.5e+02 & \textbf{4.3e+02} & \textbf{1.1e+01} & 1.9e+03 & 1.5e+02\\
\hline
24 & 5.4e+02 & 3.4e+01 & \textbf{4.9e+02} & \textbf{6.3e+00} & 5.4e+02 & 3.2e+01\\
\hline
25 & 5.5e+02 & 3.5e+01 & \textbf{4.8e+02} & \textbf{1.6e+01} & 5.5e+02 & 4.3e+01\\
\hline
26 & 2.3e+03 & 1.9e+03 & \textbf{8.6e+02} & \textbf{1.2e+02} & 2.2e+03 & 2.2e+03\\
\hline
27 & 3.3e+03 & 2.5e+02 & \textbf{5.2e+02} & \textbf{1.1e+01} & 3.3e+03 & 2.3e+02\\
\hline
28 & 4.9e+02 & 1.9e+01 & \textbf{4.8e+02} & \textbf{2.5e+01} & 5.0e+02 & 3.0e+01\\
\hline
29 & 2.4e+03 & 1.2e+03 & \textbf{4.5e+02} & \textbf{1.6e+02} & 2.2e+03 & 1.2e+03\\
\hline
30 & 6.7e+05 & 7.8e+04 & \textbf{6.2e+05} & \textbf{4.3e+04} & 6.7e+05 & 5.6e+04\\
\bottomrule
\end{tabular}
}
\caption{Statistics of results obtained by MF-CMA-ES (with PPMF) and two versions of CMA-ES (with CSA and PPMF) for CEC'2017 optimization problems in 50 dimensions after spending the budget of $10000n$ fitness evaluations.}
\label{tab:CEC17-D50}
\end{table}